\documentclass{article}

\usepackage[preprint]{neurips_2022_ml4ps}
\usepackage{amsmath}

\usepackage[utf8]{inputenc} 
\usepackage[T1]{fontenc}    
\usepackage{hyperref}       
\usepackage{url}            
\usepackage{booktabs}       
\usepackage{amsfonts}       
\usepackage{nicefrac}       
\usepackage{microtype}      
\usepackage{xcolor}         
\usepackage{graphicx}
\usepackage{wrapfig}
\usepackage[flushleft]{threeparttable}

\usepackage{natbib}
\title{Reducing Down(stream)time: Pretraining Molecular GNNs using Heterogeneous AI Accelerators}

\author{%
  Jenna A. Bilbrey \\
  Pacific Northwest National Laboratory \\
  \texttt{jenna.pope@pnnl.gov} \\
  \And
  Kristina M. Herman \\ 
  Department of Chemistry, University of Washington\\
  \texttt{kmherman@uw.edu} \\
  \And
  Henry Sprueill \\ 
  Pacific Northwest National Laboratory \\
  \texttt{henry.sprueill@pnnl.gov} \\
  \And
  Sotiris S. Xantheas \\
  Pacific Northwest National Laboratory \\
  Department of Chemistry, University of Washington\\
  \texttt{sotiris.xantheas@pnnl.gov} \\  
  \And  
  Payel Das \\
  IBM Research \\
  \texttt{daspa@us.ibm.com} \\
  \And
  Manuel Lopez Roldan \\
  Graphcore \\
  \texttt{manuelr@graphcore.ai} \\
  \And
  Mike Kraus \\ 
  Graphcore \\
  \texttt{mikek@graphcore.ai} \\
  \And
  Hatem Helal \\
  Graphcore \\
  \texttt{hatemh@graphcore.ai} \\
  \And
  Sutanay Choudhury \\
  Pacific Northwest National Laboratory \\
  \texttt{sutanay.choudhury@pnnl.gov} \\ 
}

\begin{document}
\setcitestyle{numbers}
\setcitestyle{square}

\maketitle

\begin{abstract}
The demonstrated success of transfer learning has popularized approaches that involve pretraining models from massive data sources and subsequent finetuning towards a specific task. While such approaches have become the norm in fields such as natural language processing, implementation and evaluation of transfer learning approaches for chemistry are in the early stages. In this work, we demonstrate finetuning for downstream tasks on a graph neural network (GNN) trained over a molecular database containing 2.7 million water clusters. 
The use of Graphcore IPUs as an AI accelerator for training molecular GNNs reduces training time from a reported 2.7 days on 0.5M clusters to 1.2 hours on 2.7M clusters. Finetuning the pretrained model for downstream tasks of molecular dynamics and transfer to a different potential energy surface took only 8.3 hours and 28 minutes, respectively, on a single GPU.  

\end{abstract}

\section{Introduction}

Pretraining models on massive datasets followed by finetuning towards specific downstream tasks is commonplace in natural language processing and computer vision approaches. The uptake of similar approaches for chemistry is lagging due to the limited number of large datasets and long training times involved. Training atomistic property prediction models from massive scientific datasets is a compute-intensive task, and much of the focus in recent literature has been on transformer-based models \cite{wang2019smiles, chithrananda2020chemberta}. The reduction of atomic positions to character strings via the SMILES notation has aided the generation of large datasets. For example, Wang et al.~trained the BERT architecture on 18.7M SMILES strings from the ZINC database \cite{wang2019smiles}, while Chithrananda et al.~trained a network based on the RoBERTa architecture, called ChemBERTa, on 77M SMILES strings obtained from PubChem \cite{chithrananda2020chemberta}. ChemBERTa was later shown to perform well on downstream property prediction tasks \cite{ahmad2022chemberta}. RNN-based generative models have been trained on a 1M-molecule subset of GDB-13 \cite{arus2019exploring} and a 1.6M-molecule subset of ZINC \cite{chenthamarakshan2020cogmol}. More recently, Ross et al.~developed a transformer-based encoder that uses linear attention, called Molformer, to efficiently train on $>$1000M SMILES strings, outperforming several graph- and geometry-based baselines on regression and classification tasks from benchmark datasets \cite{ross2021large}. Subsequent work has explicitly incorporated spatial properties into Molformer for training on comparatively smaller datasets \cite{wu2021molformer}.  

SMILES strings describe atom compositions and bonding configurations, but neglect information about the 3D geometry and long-range interactions, such as interactions between molecules. 
Here, we answer a benchmark challenge set forth in our prior work \cite{choudhury2020hydronet} of generating a predictive model that preserves intermolecular interactions. 
This benchmark is supported by an open-source dataset containing 4.95M unique hydrogen bonded clusters of water molecules \cite{hydrodb}. All clusters in the dataset are minima on the potential energy surface (PES) computed using the TTM2.1-F potential, making them useful for static property prediction. For example, Bilbrey et al.~\cite{bilbrey2020look} trained the SchNet neural network on a subset of 500,000 clusters and obtained a mean absolute error per water molecule of 0.002 kcal/mol on a 10,500-sample test set, which included cluster sizes outside of the range of those included in the training subset, indicating the ability of the network to extrapolate. Training on $\sim$10\% of the full benchmark dataset took 2.7 days scaled over four NVIDIA V100 GPUs \cite{bilbrey2020look}, making it impractical to train on larger subsets, much less the complete dataset, with traditional hardware. The computing power required for training neural networks on very large datasets restricts exploration of this area to researchers with ample access to GPU clusters or AI hardware accelerators \cite{guo2017survey, capra2020updated, frey2021scalable}. 

\begin{wrapfigure}{l}{0.38\textwidth}
    \centering
    \includegraphics[width=0.37\textwidth]{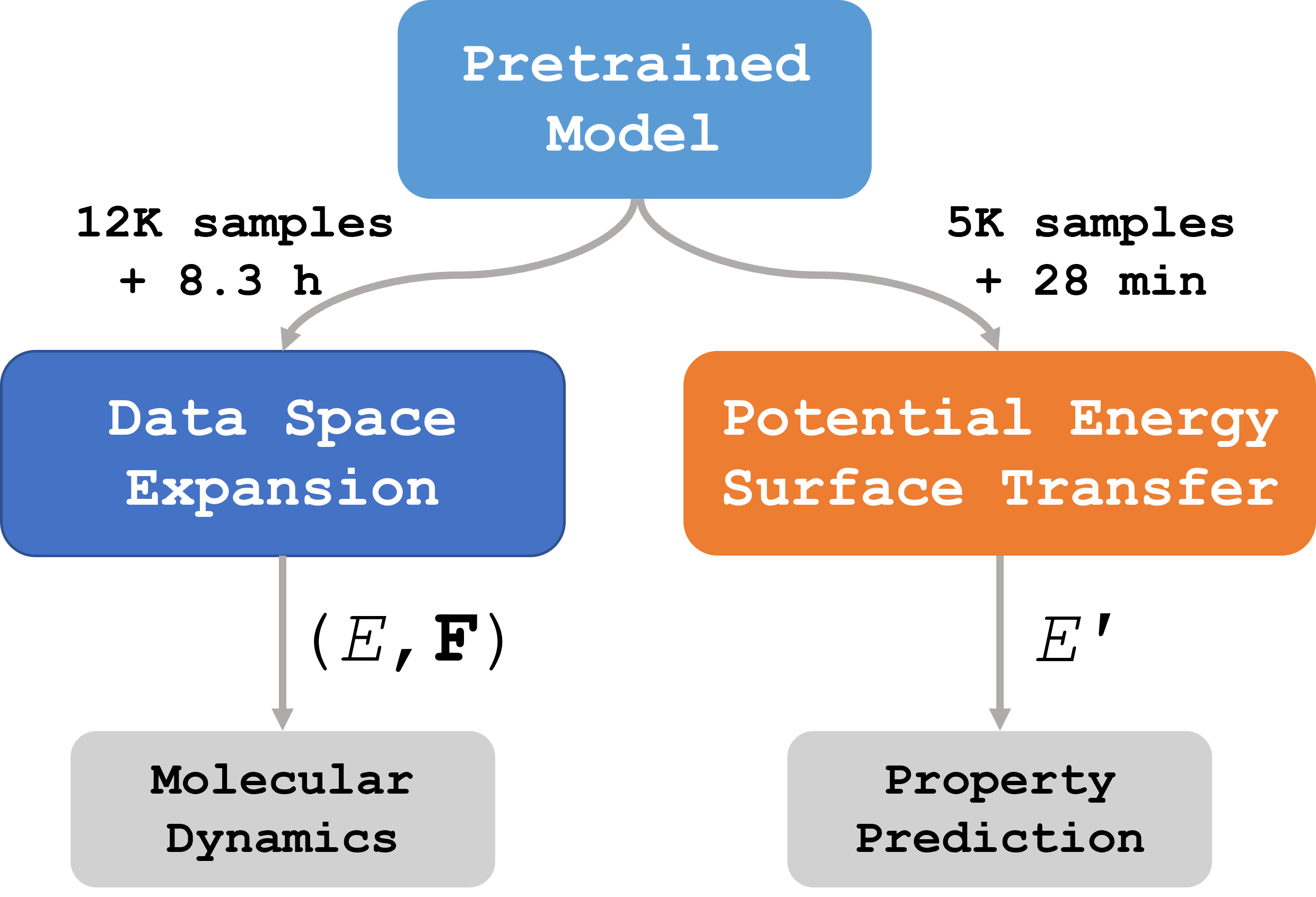}
    \caption{Downstream workflow.}
    \label{fig:workflow}
\end{wrapfigure}

In this work, we demonstrate the training and downstream inference of a graph neural network (GNN) built on a database of 2.7M molecular structures. We train the SchNet architecture \cite{schutt2018schnetpack, schutt2018schnet} on a 2.7M subset of the HydroNet benchmark dataset using Graphcore IPU hardware accelerators. We then finetune the pretrained network using much smaller datasets on a single V100 GPU for downstream tasks of molecular dynamics (MD) and property prediction. The reduced hardware and data requirements afforded by employing the pretrained model, which we make open source, open the door for experimentation by a larger number of researchers, all while decreasing energy resources consumed during experimentation. The two key factors underpinning this research are (1) the massive scale of a water cluster database that contains millions of 3D geometries and (2) the use of a novel AI accelerator that dramatically reduces training times. 

\section{Methods}

\textbf{Data Collection.} 
The dataset of water cluster minima used for pretraining was obtained from an existing database generated using Monte Carlo temperature basin-paving  (MC-TBP) simulations driven by the TTM2.1-F potential \cite{rakshit2019atlas, fanourgakis2006flexible}. 
Each cluster is associated with 3D coordinates $\mathbf{r}$ and energy $E$.
The dataset of non-minima water clusters for downstream MD via data space expansion was obtained from MD simulations performed at 260K and 300K.
Each non-minima includes atomic forces $\mathbf{F}$, $\mathbf{r}$, and $E$. To demonstrate transfer of the PES, $E$ for a subset of 5,000 minima were obtained using the MB-pol interatomic potential \cite{BabinVolodymyr2013Doa, BabinVolodymyr2014Doa}.

\textbf{Hardware Accelerators.} Accelerators designed specifically for AI/ML applications show improved processing speed, scalability, and energy efficiency, allowing faster training on larger datasets. In particular, Graphcore's IPU accelerators show a $4\times$ speedup over NVIDIA V100 GPUs for training of GNNs \cite{moe2022implementating}. Graphcore's high-level development framework PopTorch was used to implement the Pytorch Geometric (PyG) library \cite{PytorchGeometric}, which includes the SchNet framework. We trained SchNet on a dataset containing 2,726,710 water clusters, using hyperparameters reported in previous work \cite{bilbrey2020look}. 

\textbf{Finetuning.} Pretraining molecular GNNs on large datasets have been shown to improve generalization in downstream tasks \cite{hu2020pretraining}. 
Use of the PyG implementation of SchNet allows the model to be easily transferred between IPUs, GPUs, and CPUs. The saved model weights from the IPU-trained model constitute the pretrained network, which we update to obtain (1) drive MD simulations via accurate predictions of $E$ and $\textbf{F}$ on non-minima and (2) accurate predictions of $E$ on a different PES. Inference using trained models was performed on a CPU, while finetuning was performed on a single NVIDIA V100 GPU.

\textbf{Active Sampling.} We adapt an active sampling strategy when finetuning for non-minima to minimize bias when compiling the small dataset. The training set was divided into a small training subset and large reserve set. During training, the mean absolute error ($\mu_{\varepsilon}$) and standard deviation of errors ($\sigma_{\varepsilon}$) in $\textbf{F}$ were computed over the validation set. Then, $\varepsilon_s$ was calculated for a subset of reserve samples. A sample was moved from the reserve set to the training subset if $1- \text
{erf}\left((\varepsilon_s-\mu_{\varepsilon})/\sigma_{\varepsilon}\right)<p_{\text{tol}}$, where $p_{\text{tol}}$ is a chosen tolerance and $\text{erf}(x)$ is the Gaussian error function.

\section{Results and discussion}

We present performance between three model variations: SchNet trained on 2.7M water cluster minima (\textsc{pretrained}), the pretrained model further finetuned using a much smaller dataset (\textsc{finetuned}), and a SchNet model trained from scratch using only the smaller dataset (\textsc{scratch}) -- on two downstream tasks -- data space expansion and PES transfer.

\textbf{Model Pretraining.} The SchNet model was trained on 2.7M water clusters of size $N$=3--25 with a 0.8:0.1:0.1 train-validation-test split. 
Training took 4.2, 2.2, or 1.2 hours scaled across 16, 32, or 64 IPUs, respectively. The validation loss increased with the number of IPUs to 0.0017, 0.0020, and 0.0030, respectively; therefore, we used the model trained over 16 IPUs to finetune for subsequent downstream tasks. This model showed a similarly low test-set error of 0.0018 kcal/mol.

\begin{table}[b]
\centering
  \caption{
  Test-set accuracy, reported as mean absolute error (MAE) in $E_{\mathrm{H_2O}}$ (kcal/mol), $\mathit{F}_{\mathrm{mag}}$ (kcal/mol/\AA), and $\mathit{F}_{\mathrm{ang}}$. 
  Test sets were derived from the specified dataset.}
  \label{static-error}
  \centering
  \begin{tabular}{lllllll}
    \toprule
    Hardware	&	Dataset	&	Initialization	&	$N_{\mathrm{train}}$	&	$E_{\mathrm{H_2O}}$	&		$\mathit{F}_{\mathrm{mag}}$	&	$\mathit{F}_{\mathrm{ang}}$	\\
    \midrule
    IPU	&	minima  &	scratch	   &	2.7M	&	0.0018	&	21.89	&	0.501	\\
    GPU	&	non-minima	    &	scratch	   &	11.8K	&	0.3548	&	18.76	&	0.238	\\
    GPU	&	non-minima	    &	pretrain   &	11.6K	&	0.1316	&	8.94	&	0.098	\\
    \bottomrule
  \end{tabular}
\end{table}

\begin{figure}[t]
  \centering
  \includegraphics[width=\textwidth]{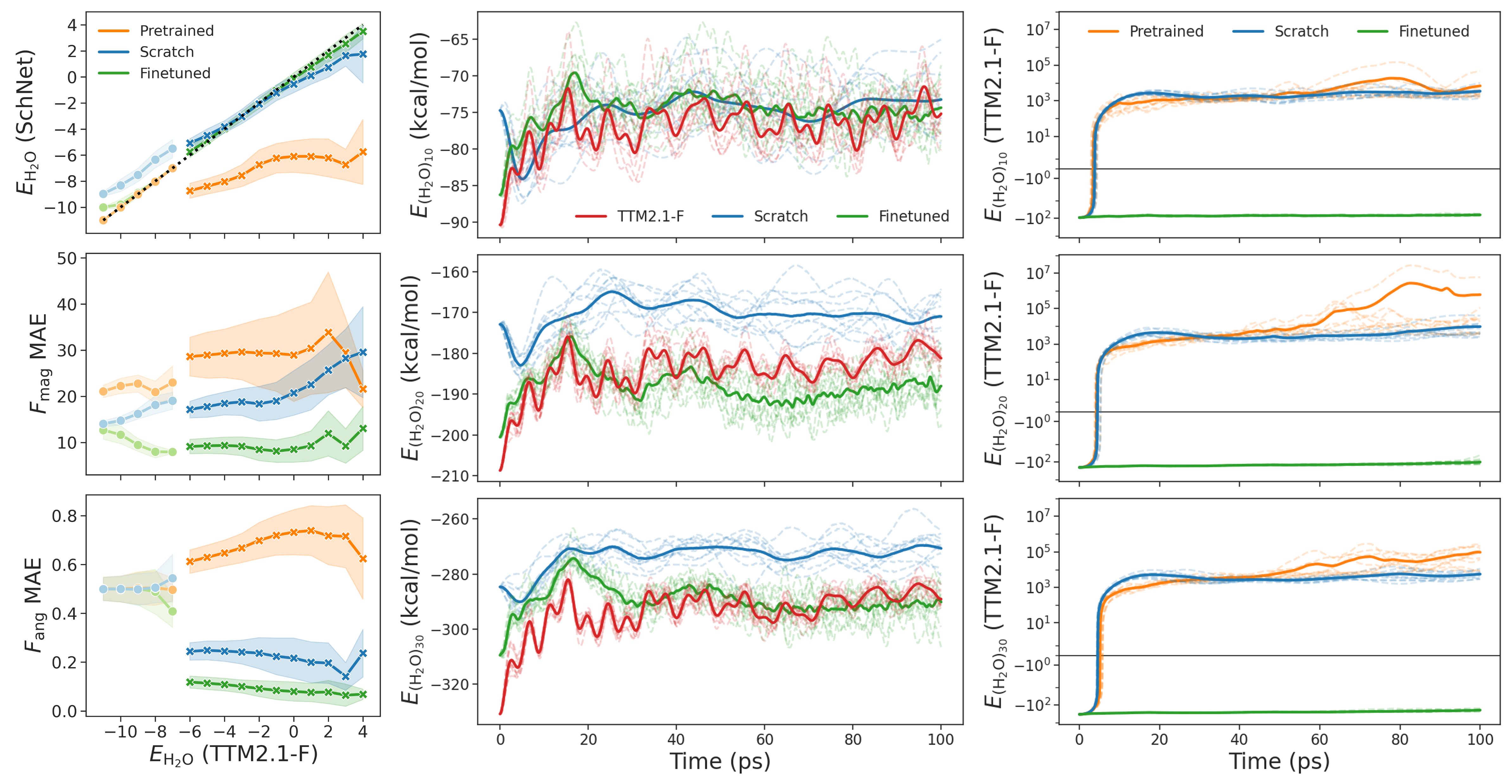}
  \caption{
  (Left) Static predictions of $E_{\mathrm{H_2O}}$, $\mathit{F}_{\mathrm{mag}}$, and $\mathit{F}_{\mathrm{ang}}$
  on minima ($\bullet$) and non-minima ($\pmb{\times}$) test sets. (Center) MD using the TTM2.1-F potential and NNPs for water clusters of size $N$=10, 20, 30
  . (Right) Static TTM2.1-F predictions on MD trajectories generated by the NNPs.}
  \label{fig:finetune}
\end{figure}

\textbf{Data Space Expansion.} 
MD simulations explore non-minima on the PES and can be driven by neural network potentials (NNPs). The forces $\mathbf{F}$ used to drive atomic motions are obtained as the negative of the NNP gradient. A force term is typically added to the loss function to improve prediction accuracy \cite{chmiela2017machine}. 
Training without the force term, as was done for the pretrained model, leads to poor prediction of  $\mathbf{F}$, even through the accuracy in $E$ predictions on minima configurations is high. Moreover, the pretrained model produces poor $E$ predictions on non-minima, necessitating the need for the data space covered by the model to be expanded. Finetuning the pretrained model on a much smaller subset of non-minima ($>$1\% of the minima dataset) and including a force term in the loss produces a NNP with good predictions of non-minima $\mathbf{F}$ and $E$. Because of the roughly three-order-of-magnitude decrease in the size of the training set when using the pretrained model, training was accomplished on a single NVIDIA V100 GPU in 8.3 hours. 

Table \ref{static-error} shows the test-set accuracy of the pretrained model and models trained on non-minima with and without pretraining. 
The test set corresponds to the specific training set, i.e., the pretrained model is tested on minima, while the models trained on non-minima are tested on non-minima. 
Following Chmiela et al. \cite{chmiela2017machine}, we quantify the topological accuracy of atomic force predictions by the magnitude error 
$F_{\mathrm{mag}}=\lVert \hat{\textbf{F}} \rVert-\lVert \textbf{F} \rVert$,
which describes the extent to which the slope of the predicted and reference PES differ, and the angular error 
$F_{\mathrm{ang}}=\textrm{cos}^{-1}(\hat{\textbf{F}}/\lVert \hat{\textbf{F}} \rVert \cdot \textbf{F}/\lVert \textbf{F} \rVert)/\pi$,
which describes the orientation of the predicted force direction relative to the reference force direction and ranges between 0 (aligned) and 1 (inverted).
Figure \ref{fig:finetune} shows static errors for the three models on minima and non-minima.  
Training from scratch on non-minima greatly improves static predictions of $E$ and provides some improvement in $\textbf{F}$ predictions, while finetuning from the pretrained model greatly improves predictions of $\textbf{F}$ and $E$ for both minima and non-minima. 

We then performed MD simulations driven by the NNPs (Fig.~\ref{fig:finetune}). Berendsen NVT dynamics of three clusters ($N=10, 20, 30$) at 300K were simulated 10 times each with different random seeds. The mean (bold lines) and individual simulations (dashed lines) are shown for TTM2.1-F and the NNPs trained on non-minima. 
MD simulations using the pretrained model are not shown, as the predicted energies were several orders of magnitude below those from TTM2.1-F. 
The mean $E$ of the finetuned model aligns for all cluster sizes, while that of the model without pretraining aligns only for $N=10$. We then calculated the TTM2.1-F $E$ on each point in the NNP-generated trajectories to validate the resulting molecular structures. Notably, only the finetuned model produced valid dynamics ($E<0$ kcal/mol). Though the model without pretraining predicted $E$ of similar magnitude to TTM2.1-F, the generated structures were calculated by TTM2.1-F to be highly unstable. In fact, the model trained from scratch did not outperform even the pretrained model in generating stable dynamics.

\begin{wrapfigure}{l}{0.5\textwidth}
  \centering
  \includegraphics[width=0.48\textwidth]{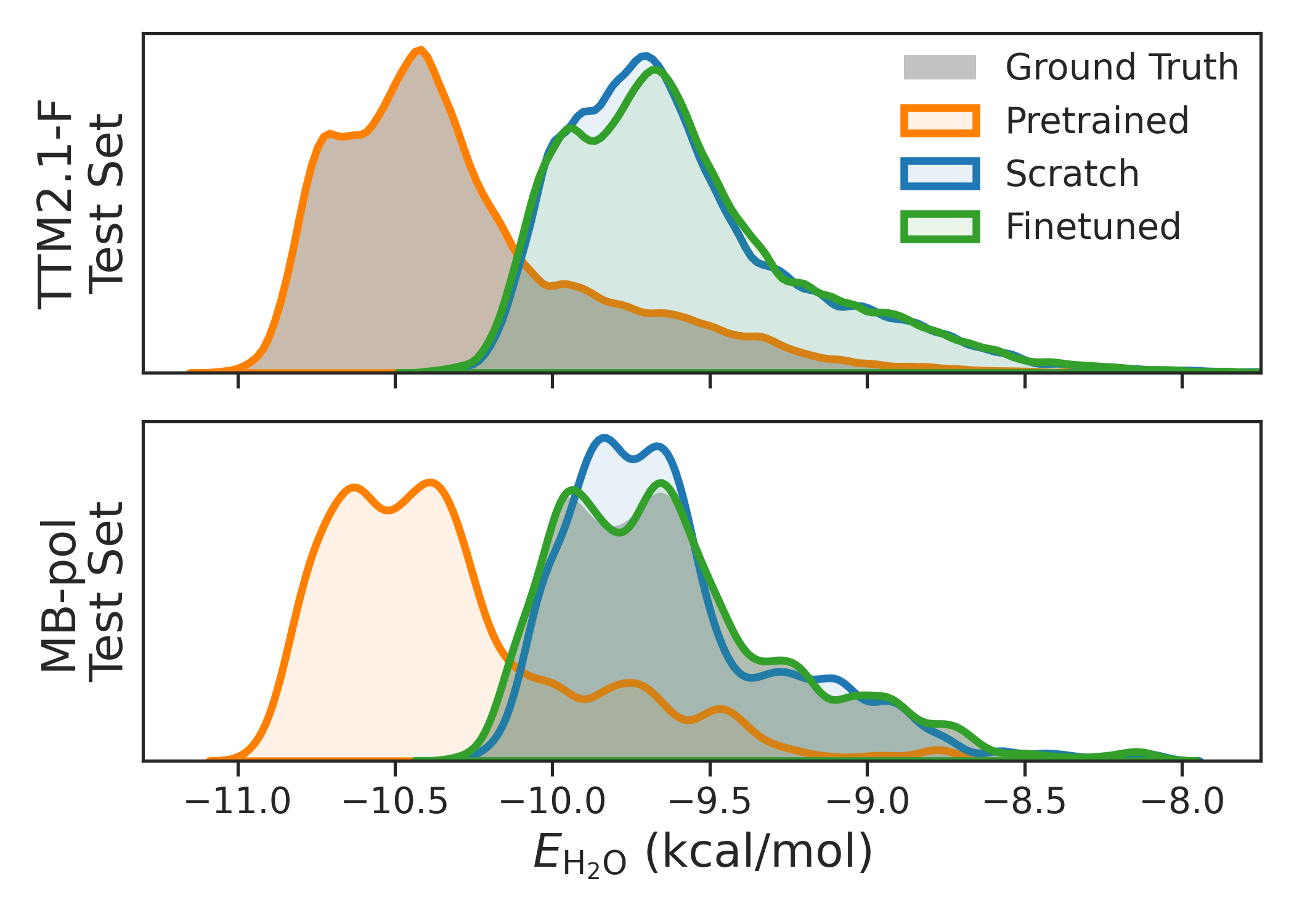}
  \caption{
  Predicted $E_{\mathrm{H_2O}}$ distributions on test sets computed with the TTM2.1-F (top) and MB-pol (bottom) potentials.
  }
  \label{fig:transfer}
\end{wrapfigure}

\textbf{PES Transfer.} The PES of a physical system will have a different representation depending on the method of calculation, each of which can have different associated computational costs. 
Classical many-body force fields, such as TTM2.1-F, are fast enough to generate large datasets, though at reduced accuracy, while \textit{ab initio} methods are highly accurate but prohibitively expensive, making the generation of comparably sized training sets intractable.
The PES approximated by a model can be amended by updating a model trained on a large dataset collected by a specific method (here, using the TTM2.1-F potential) using a small dataset collected by an alternative method (here, the MB-pol potential).
We perform such PES transfer by finetuning the pretrained model on 5,000 samples with $E$ computed with the MB-pol potential. Because of the drastically reduced size of the dataset, the model was trained in under 28 minutes on 1 NVIDIA V100 GPU. Figure \ref{fig:transfer} shows $E_{\mathrm{H_2O}}$ distributions for the TTM2.1-F and MB-pol test sets, with errors shown in Table \ref{transfer}. A shift in $E_{\mathrm{H_2O}}$ towards higher values is seen for the MB-pol potential and is reproduced for both models trained on MB-pol data. However, the finetuned model showed a $\sim$17\% lower error and more accurately reproduced the $E_{\mathrm{H_2O}}$ distribution. 

\begin{table}
\centering
  \caption{Test-set errors in $E_{\mathrm{H_2O}}$ (kcal/mol).
}
  \label{transfer}
  \centering
  \begin{tabular}{llllll}
    \toprule
    Hardware	&	Initialization	&	Train Set 	&	Test Set 	&	MAE	&	RMSE	\\
    \midrule
IPU	&	scratch	&	TTM2.1-F	&	TTM2.1-F	&	0.0018	&	0.0032	\\
GPU	&	scratch	&	MB-pol	&	TTM2.1-F	&	0.6973	&	0.7009	\\
GPU	&	pretrain	&	MB-pol	&	TTM2.1-F	&	0.7033	&	0.7071	\\
IPU	&	scratch	&	TTM2.1-F	&	MB-pol	&	0.7071	&	0.7112	\\
GPU	&	scratch	&	MB-pol	&	MB-pol	&	0.0719	&	0.0924	\\
GPU	&	pretrain	&	MB-pol	&	MB-pol	&	0.0122	&	0.0158	\\
    \bottomrule
  \end{tabular}
\end{table}

\section{Conclusions}
We demonstrate that pretraining with a very large dataset of molecular structures improves downstream tasks such as driving MD simulations as well as transfer learning to a different PES. This paper is the first to demonstrate the effectiveness of the fine-grained parallelism of the Graphcore IPU architecture for training molecular GNNs. Initial training was accomplished over 16 IPUs in 4.2 hours, while finetuning with a small set of non-minima was accomplished in 8.3 hours on a single NVIDIA V100 GPU and transfer learning with a very small set of minima computed by a different method was accomplished in only 28 minutes. Pretaining was shown to decrease the amount of data and time required, as well as reduce hardware requirements, increasing training throughput and improving accessibility to researchers with limited resources. The pretrained model was not finetuned for downstream tasks on molecules other than water; moving to a separate area of chemical space, for example, organic small molecules, could be accomplished by following our workflow, i.e., training on an open dataset of minima and finetuning for the desired downstream task on a small bespoke dataset. 

\section*{Data Availability}
The full database of water cluster minima computed with the TTM2.1-F potential is available for download at \url{https://data.pnnl.gov/group/nodes/dataset/33224}. The preprocessed databases for training, including the database of nonminima computed with the TTM2.1-F potential and the database of minima computed with the MB-pol potential, dataset split files, trained model state dictionaries, ASE databases used for MD simulations, and results of data space expansion and PES transfer analyses are available for download at \url{https://data.pnnl.gov/group/nodes/dataset/33283}.

\section*{Code Availability}
The codebase used to finetune the pretrained model for the tasks of data space expansion and PES transfer, along with hyperparameters used in this work, is available at \url{https://github.com/pnnl/downstream_mol_gnn}.

\section*{Impact statement}
IPUs show reduced energy consumption compared with GPUs when training neural networks. In addition, pretraining reduces data and hardware requirements, leading to reduced energy consumption and increased accessibility to researchers with limited resources.

\begin{ack}
The authors thank Dr.~Logan Ward for fruitful discussions on neural network potentials. 
J.A.B., K.M.H., H.S., S.S.X., and S.C. were supported by the DOE Exascale Computing Project, ExaLearn Co-design Center. The research was performed using resources available through Research Computing at Pacific Northwest National Laboratory (PNNL). PNNL is operated by Battelle for the U.S. Department of Energy under Contract DE-AC05-76RL01830. 
\end{ack}

{
\small

\bibliography{bibliography}
}

\end{document}